\pdfoutput=1
\documentclass[10pt,twocolumn,letterpaper]{article}

\usepackage[iccvfinal]{iccv}
\usepackage{multirow}
\usepackage{listings}
\usepackage{lstautogobble}
\definecolor{bluekeywords}{rgb}{0.13, 0.13, 1}
\definecolor{greencomments}{rgb}{0, 0.5, 0}
\definecolor{redstrings}{rgb}{0.9, 0, 0}
\definecolor{graynumbers}{rgb}{0.5, 0.5, 0.5}
\definecolor{iccvblue}{rgb}{0.21,0.49,0.74}
\usepackage[pagebackref,breaklinks,colorlinks,allcolors=iccvblue]{hyperref}

\def\paperID{7242} 
\def\confName{ICCV}
\def\confYear{2025}
\definecolor{zrj}{rgb}{0.2,0.6,1.0}

\definecolor{Qing}{rgb}{0,0,0}
\newcommand{\qing}[1]{{\color{Qing} {#1}}}
\title{DNF-Intrinsic: Deterministic Noise-Free Diffusion for Indoor Inverse Rendering}

\author{
    Rongjia Zheng$^1$\quad  
    Qing Zhang$^{1,3}$\thanks{Corresponding author.}\quad  
    Chengjiang Long$^{2}$\quad Wei-Shi Zheng$^{1,3}$ \\
   \small $^1$ Sun Yat-sen University, China\quad \small $^2$ Meta Reality Labs, USA\\
    \small $^3$ Key Laboratory of Machine Intelligence and Advanced Computing, Ministry of Education, China\\
}

\makeatletter
\apptocmd\@maketitle{{\myfigure{}\par}}{}{}

\begin{document}
\newcommand{\myfigure}{
    \centering
    \vspace{-2em}
    \includegraphics[width=0.99\textwidth]{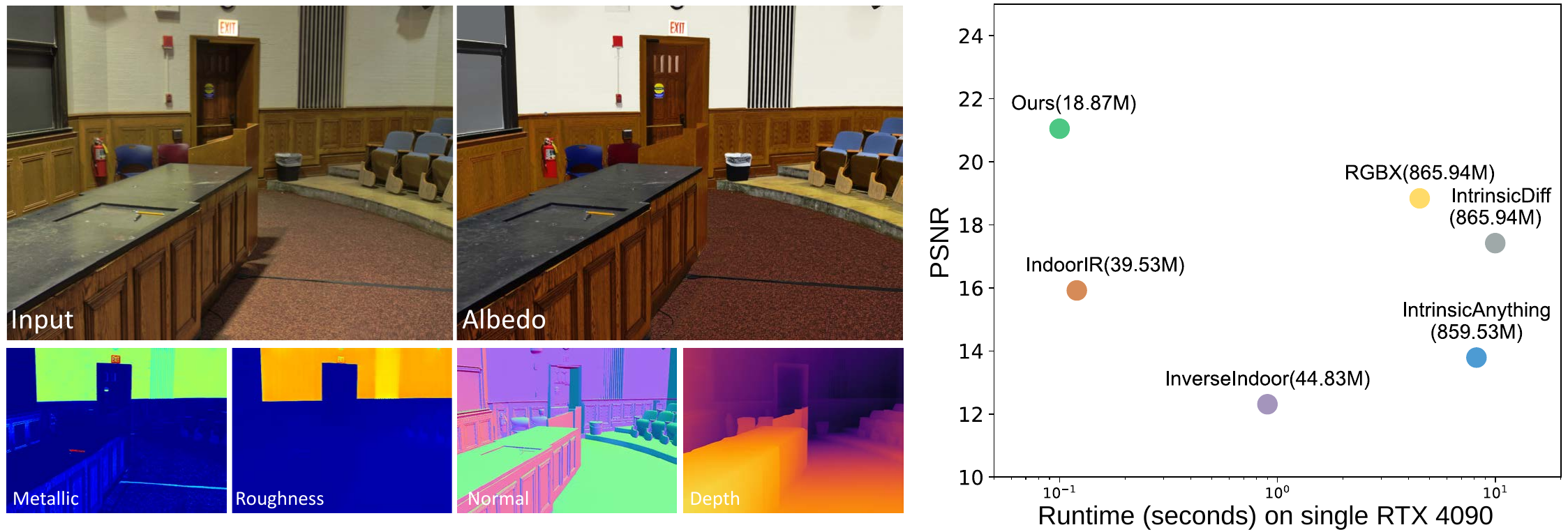}\\
  \vspace{-0.5em}
\captionsetup{type=figure}\caption{\textbf{Performance of our method on indoor inverse rendering.} By fine-tuning a pre-trained diffusion model on InteriorVerse ~\cite{zhu2022learning}, a synthetic indoor scene dataset, we achieve state-of-the-art indoor inverse rendering that can robustly recover albedo, metallic, roughness, normal, and depth from a single RGB indoor image. Taking albedo estimation as an example, as shown on the right, our method not only obtains significantly better results (21.05 in PSNR) than existing methods, \eg, InverseIndoor~\cite{li2020inverse}, IndoorIR~\cite{zhu2022learning}, IntrinsicAnything~\cite{chen2024intrinsicanything}, IntrinsicDiff~\cite{kocsis2024intrinsic}, and RGBX~\cite{zeng2024rgb}, but also enjoys the fastest inference speed (0.1 seconds with a single step) and fewest trainable parameters (18.87M). Note, the PSNR and runtime performance on the right are evaluated on the test set of the InteriorVerse dataset.}  
\label{fig:teaser}
\vspace{12pt}
}
\maketitle
\begin{abstract}
Recent methods have shown that pre-trained diffusion models can be fine-tuned to enable generative inverse rendering by learning image-conditioned noise-to-intrinsic mapping. Despite their remarkable progress, they struggle to robustly produce high-quality results as the noise-to-intrinsic paradigm essentially utilizes noisy images with deteriorated structure and appearance for intrinsic prediction, while it is common knowledge that structure and appearance information in an image are crucial for inverse rendering. To address this issue, we present DNF-Intrinsic, a robust yet efficient inverse rendering approach fine-tuned from a pre-trained diffusion model, where we propose to take the source image rather than Gaussian noise as input to directly predict deterministic intrinsic properties via flow matching. Moreover, we design a generative renderer to constrain that the predicted intrinsic properties are physically faithful to the source image. Experiments on both synthetic and real-world datasets show that our method clearly outperforms existing state-of-the-art methods. Our code is available at \href{https://github.com/OnlyZZZZ/DNF-Intrinsic}{\textcolor{red}{https://github.com/OnlyZZZZ/DNF-Intrinsic}}.
\end{abstract}
\section{Introduction}
\label{sec:intro}
Inverse rendering has long been a fundamental task in computer vision and graphics since originated by Barrow and Tenenbaum \cite{barrow1978recovering}. It aims to reverse the rendering process by estimating the scene material (\eg, albedo, metallic, and roughness), geometry (\eg, normal and depth), and lighting from a single image. This task was widely studied as acquiring these intrinsic properties allows a wide range of applications including AR/VR \cite{liang2024photorealistic}, segmentation and tracking \cite{baslamisli2018joint,ost2024inverse}, relighting and material editing \cite{yeh2022photoscene}, as well as robotics and autonomous driving \cite{chaudhury2024shape}.

Currently, inverse rendering remains a challenging open problem due to the following reasons. First, the problem is highly ill-posed, especially when only a single image with complex appearance effects (\eg, inter-reflection and cast shadows) is given. Second, there is a lack of large-scale real-world datasets with ground-truth intrinsic properties in this field, which makes it difficult to achieve inverse rendering of real-world scenes in a data-driven manner.

Early methods often resort to hand-crafted priors to reduce the ambiguity of the problem \cite{finlayson2004intrinsic,barron2012shape,barron2014shape}. However, as these priors do not always hold in real-world images, they may not work well for complex scenes. Benefiting from the emergence of large-scale synthetic datasets~\cite{li2020openrooms,zhu2022learning,DIODE}, learning-based methods~\cite{du2023generative,li2020inverse,zhu2022learning,zhu2022irisformer,li2022physically,li2022phyir,li2023inverse,li2023multi} casting inverse rendering as fully supervised image-to-image translation later became mainstream, but they often do not generalize well to real-world images.

A recent trend that has received considerable attention is to leverage the strong image prior of pre-trained diffusion models to achieve non-deterministic inverse rendering in a generative noise-to-intrinsic fashion. Following this paradigm, some methods propose to tackle inverse rendering by fine-tuning pre-trained diffusion models using synthetic data \cite{kocsis2024intrinsic,chen2024intrinsicanything,luo2024intrinsicdiffusion,zeng2024rgb}. However, they basically have the following three limitations. First, they cannot robustly produce high-quality results because the use of noise-to-intrinsic paradigm makes it impossible for them to exploit the complete image structure and appearance information for intrinsic prediction. Second, their inference speed is generally low, as a large number of denoising steps are usually required to map random noise to intrinsic properties. Finally, due to the lack of explicit constraint on image reconstruction from the predicted intrinsic properties, they may fail to achieve physically convincing inverse rendering. 

To address these limitations, we present DNF-Intrinsic, a diffusion-based inverse rendering approach that allows for robust, efficient, and physically convincing single-image inverse rendering of indoor scenes (see Figure~\ref{fig:teaser}). Instead of learning image-conditioned noise-to-intrinsic mapping like previous methods, we advocate for image-to-intrinsic mapping \footnote{As verified by \cite{bansal2023cold,martingarcia2024diffusione2eft} and our experiments, the prior in pre-trained diffusion model still holds when the model is fine-tuned with noise-free image-to-target mapping.}, where we take the source image rather than Gaussian noise as input to predict intrinsic properties in a deterministic manner using flow matching. This not only allows our method to obtain better intrinsic prediction performance by taking full advantage of the visual information in the source image, but also helps avoid the high computational cost arising from a large number of denoising steps. A generative renderer is further designed to achieve physically meaningful inverse rendering by explicitly constraining that the source image can be reconstructed from the predicted intrinsic properties. In summary, our main contributions are:
\begin{itemize}[leftmargin=2em]
\setlength\itemsep{0.5em}
\item We present a deterministic noise-free diffusion-based approach that enables robust yet efficient single-image inverse rendering of indoor scenes.

\item We show that image-to-intrinsic mapping rather than image-conditioned noise-to-intrinsic mapping is a more effective choice for diffusion-based inverse rendering. Besides, we design a generative renderer to constrain the predicted intrinsic properties to be physically more reliable. 

\item Extensive experiments show that our method significantly outperforms state-of-the-art approaches on both synthetic and real-world datasets.
\end{itemize}

\section{Related Work}
\label{sec:related}
\begin{figure*}[t]
\centering
\includegraphics[width=\textwidth]{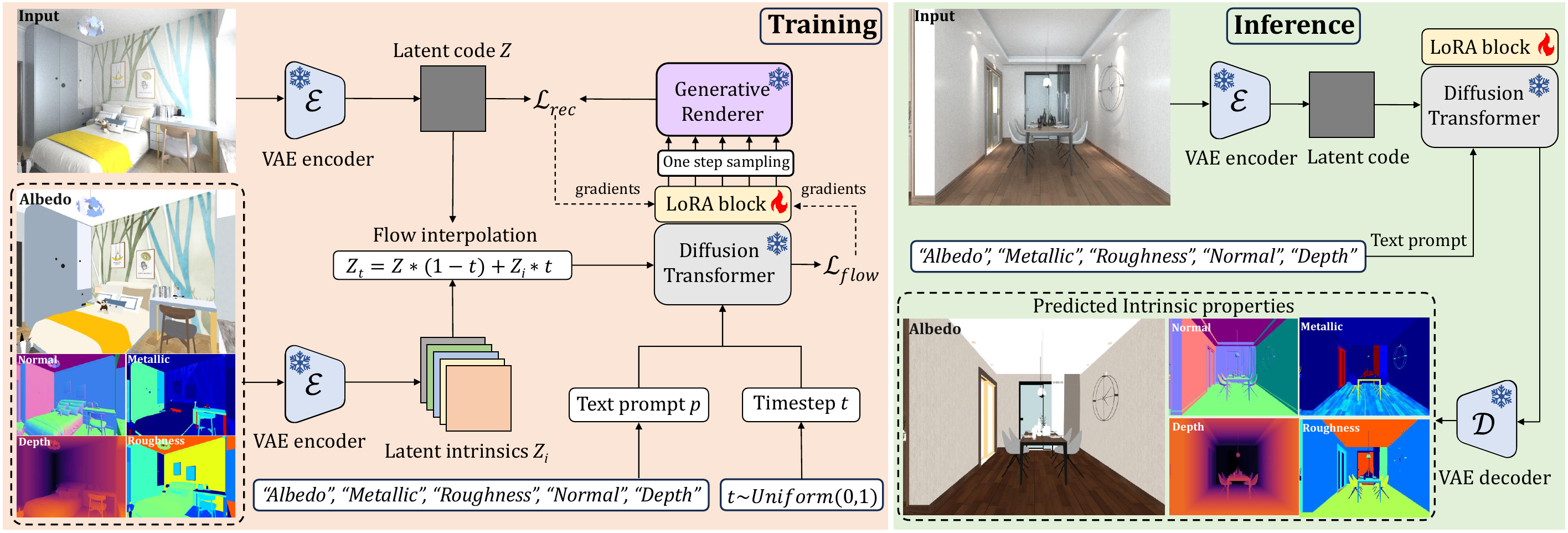} \\
\caption{
\textbf{Overview of our method.} Given a single indoor image, we aim to recover albedo, metallic, roughness, normal, and depth by learning deterministic image-to-intrinsic mappings triggered by text prompts. During training, instead of starting from noise as done in previous diffusion-based methods, we employ a pre-trained VAE encoder $\mathcal{E}$ to initialize the latent code $Z$ of the input image as the starting point of the flow trajectory. Then, the noised latent $Z_t$ at timestep $t$ is obtained by interpolation between $Z$ and the latent intrinsic $Z_i$. Supervised by $\mathcal{L}_{flow}$, each noise latent and the corresponding text prompt are sent into a pre-trained diffusion transformer (DiT) with a trainable LoRA block to predict the flow velocity toward each intrinsic space. In addition, based on the one step sampling of all intrinsic properties, we develop a generative renderer to accordingly design a reconstruction loss $\mathcal{L}_{rec}$ to provide supervision from the input image. During inference, starting from the latent code $Z$, we traverse each intrinsic flow triggered by its corresponding text prompt to obtain the predicted latent intrinsic and recover the final intrinsic properties using a VAE decoder $\mathcal{D}$.}

\label{figure_framework}
\end{figure*}


\noindent \textbf{Optimization/learning-based approaches.} Early works in this field focus on tackling intrinsic image decomposition (IID), a sub-problem of inverse rendering that aims to separate an image into reflectance and shading. Conventional IID methods mainly work by formulating optimization based on hand-crafted priors \cite{barron2012shape,finlayson2004intrinsic}, while subsequent IID methods are mostly learning-based~\cite{das2022pie,das2023idtransformer,careaga2023intrinsic,careaga2024colorful,zhang2021unsupervised,careaga2023Harmonization}, due to the availability of various datasets~\cite{bell2014intrinsic,kovacs2017shading}. In addition to IID, other sub-tasks of inverse rendering such as material/lighting estimation \cite{gardner2019deep,legendre2019deeplight} and geometry (\eg, depth and normal) prediction \cite{bae2024rethinking,depthanything,depth_anything_v2} also receive considerable research attention. However, these methods can only estimate one or two of the intrinsic properties. Although there are inverse rendering methods that enable joint estimation of several intrinsic properties, they are mostly tailored for single object~\cite{verbin2024eclipse,enyo2024diffusion,chung2024differentiable} or require multi-view images as input~\cite{li2023multi,boss2021nerd,munkberg2022extracting}. By training on large-scale synthetic datasets~\cite{li2020openrooms,zhu2022learning,sengupta2019neural,roberts2021hypersim}, several works achieve single-image inverse rendering of indoor scenes~\cite{li2022phyir,sengupta2019neural,li2020inverse,zhu2022learning,zhu2022irisformer,wang2021learning}, but their performance may severely deteriorate on real-world images due to the inherent domain gap issue.   

\vspace{0.5em}
\noindent \textbf{Diffusion-based approaches.} A recent research trend in this field is to leverage the powerful image prior learned by pre-trained diffusion models for tasks including geometry prediction \cite{ke2024repurposing,fu2025geowizard}, material estimation \cite{kocsis2024intrinsic}, and inverse rendering \cite{luo2024intrinsicdiffusion,zeng2024rgb,chen2024intrinsicanything}. However, these methods mostly adopt a similar image-conditioned noise-to-target paradigm, which is highly sensitive to random noise initialization and introduces undesirable detail distortions. As shown by Tables \ref{table:synthetic} and \ref{table:real_com} as well as Figure~\ref{figure_real_albedo}, they fail to enable robust performance and typically correspond to low inference speed due to the requirement of a large number of diffusion denoising steps (see Figures~\ref{fig:teaser} and~\ref{figure_steps}).

\section{Method}
\label{sec:method}

 Given a single RGB indoor image, the goal of our method is to recover its multiple intrinsic properties including albedo, metallic, roughness, normal, and depth via a single model fine-tuned from a pre-trained diffusion model. Figure~\ref{figure_framework} shows the overview of our method.

\subsection{Problem Formulation}
Unlike text-to-image generation, inverse rendering is a deterministic image-to-image task. To leverage the powerful diffusion prior, most previous diffusion-based inverse rendering methods choose to follow text-to-image methods that formulate the task as a noise-to-intrinsic denoising diffusion problem conditioned on the input image. Taking albedo estimation as an example, the pre-trained diffusion model is usually trained to learn the conditional distribution $D(A|I)$ over albedo $A\sim \mathbb{R}^{ W \times H \times 3}$, where the condition $I\sim \mathbb{R}^{ W \times H \times 3}$ is an input image. The training objective of the diffusion model $\epsilon_{\theta}$ is formulated as:
\begin{equation} \label{eq1}
    x_t = \sqrt{\bar{\alpha}_t}A + \sqrt{1 - \bar{\alpha}_t} \epsilon, t \in \{0, 1, \dots, T\},
\end{equation}
\begin{equation}
\mathcal{L} = \mathbb{E}_{t,A,\epsilon} \left\| \epsilon_{\theta}(x_t, t, I) - \epsilon \right\|_2^2, 
\end{equation}
where $\bar{\alpha}_t$ refers to noise schedule, $x_t$ is the noised image at timestep $t$ and $\epsilon \sim \mathcal{N}(0, \mathbf{I})$ represents Gaussian noise. Next, the backward denoising process gradually remove the noise in $x_t$ to get $x_{t-1}$:
\begin{equation}
d(x_t) = \sqrt{1 - \bar{\alpha}_{t-1}} - \tau^2 \epsilon_{\theta}, \label{eq_2}
\end{equation} 
\begin{equation}
x_{t-1} = \sqrt{\bar{\alpha}_{t-1}}(\frac{x_t - \sqrt{1 - \bar{\alpha}_t}\epsilon_{\theta}}{\sqrt{\bar{\alpha}_t}}) + d(x_t) + \tau \epsilon,\label{eq_3}
\end{equation}
where $\tau$ is a parameter to control the amount of injected noise, and $\epsilon_{\theta} = \epsilon_{\theta}(x_t, t, I)$ is the predicted noise. During inference, it starts with a pure Gaussian noise image $x_{T}$ and gradually remove noise from $t \in \{T\,\dots,1,0 \}$ via Eq. \eqref{eq_3}, eventually leading to a clean albedo prediction.

\begin{figure*}[t]
\centering
\includegraphics[width=\textwidth]{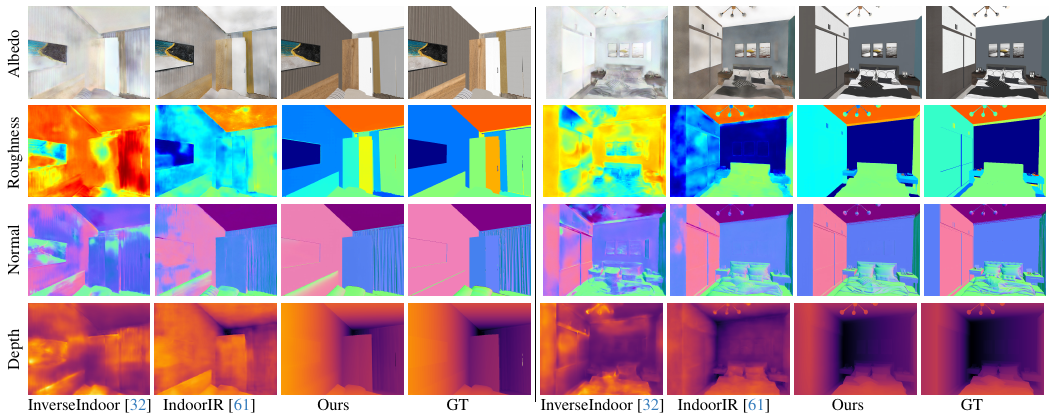} \\
\caption{\textbf{Qualitative comparison of inverse rendering on the synthetic InteriorVerse dataset~\cite{zhu2022learning}.} Although both our method and IndoorIR~\cite{zhu2022learning} enable metallic estimation, we here do not compare on metallic as InverseIndoor~\cite{li2020inverse} cannot predict it. Please see the supplementary material for more results and comparison on metallic estimation.}

\label{fig:syn_comp}
\end{figure*}
\vspace{0.5em}
\noindent \textbf{Our motivation.} Although the aforementioned image-conditioned noise-to-intrinsic paradigm has demonstrated promising results on material \cite{kocsis2024intrinsic,chen2024intrinsicanything,zeng2024rgb,luo2024intrinsicdiffusion} and geometry estimation \cite{ke2024repurposing,fu2025geowizard,ye2024stablenormal}, there are still several issues to be addressed. First of all, this paradigm learns to operate on noisy images with deteriorated structure and appearance, which inevitably harms the overall performance as structure and appearance information are known to be crucial for inverse rendering \cite{cheng2023structure,fan2018revisiting,barron2014shape}. Second, the inference speed of this paradigm is mostly low due to the inherent difficulty of mapping noise to target and the curved sampling trajectory defined by Eq. \eqref{eq_3}. Third, methods following this paradigm struggle to produce physically plausible high-quality intrinsic properties because they are often trained without explicitly enforcing that the predicted intrinsic properties are able to reconstruct the input image. The existence of the above issues motivates us to develop a diffusion-based inverse rendering method that learns image-to-intrinsic mapping taking the source image as input to predict reconstruction-constrained high-quality intrinsic properties in a deterministic and more efficient manner. 

\subsection{Deterministic Image-to-Intrinsic Diffusion}
To enable more robust and efficient inverse rendering, we propose to learn a deterministic image-to-intrinsic mapping via flow matching \cite{lipman2022flow,liu2022flow}, rather than DDPM \cite{ho2020denoising} based image-conditioned noise-to-intrinsic mapping like previous methods. Specifically, we use an ordinary differential equation (ODE) model to transfer the latent code of input image $Z \sim \pi_I$ to the target latent intrinsic $Z_p \sim \pi_p$ via a short and straight trajectory, enabling prediction with only a small number of sampling steps. During training, the noised latent $Z_t$ is obtained by interpolation:
\begin{equation}
    Z_t = Z*(1 - t) + Z_p*t, t \in [0, 1],
\end{equation}
where $t \in [0, 1]$ is a normalized timestep.
Then, a neural network $\mu_{\theta}$ is employed to learn the flow velocity along the path between the input image and each single intrinsic property using the loss below:
\begin{equation}
\mathcal{L}_{single} = \mathbb{E}_{t,Z,Z_p} \left\| \mu_{\theta}(Z_t, t) - (Z_{p}-Z) \right\|_2^2. \label{eq:5}
\end{equation}

\begin{table*}[tb]
    \vspace{-2mm}
    \centering
    \resizebox{\linewidth}{!}{
    \begin{tabular}{lcccccccccccc}
    \toprule[1pt]
    \multirow{2}{*}{Method}  & \multicolumn{3}{c}{Albedo} & \multicolumn{3}{c}{Metallic} & \multicolumn{3}{c}{Roughness} & \multicolumn{1}{c}{Normal} & \multicolumn{1}{c}{Depth}\\
    &\multicolumn{1}{c}{PSNR$\uparrow$} &\multicolumn{1}{c}{SSIM$\uparrow$} &\multicolumn{1}{c}{LPIPS$\downarrow$} & \multicolumn{1}{c}{PSNR$\uparrow$} &\multicolumn{1}{c}{SSIM$\uparrow$} & \multicolumn{1}{c}{LPIPS$\downarrow$} & \multicolumn{1}{c}{PSNR$\uparrow$} &\multicolumn{1}{c}{SSIM$\uparrow$} & \multicolumn{1}{c}{LPIPS$\downarrow$} & \multicolumn{1}{c}{AE$\downarrow$} & \multicolumn{1}{c}{AMRE$\downarrow$}
    \\ \midrule
    InverseIndoor~\cite{li2020inverse} & 12.31 & 0.68 & 0.52 & - & - & - & 11.76 & 0.53 & 0.45 & 26.07 & 0.25\\
    IndoorIR~\cite{zhu2022learning} & 15.92 & 0.78 & 0.34 & 16.72 & 0.60 & 0.27 & 16.13 & 0.68 & 0.29 & 15.41 & 0.13\\
    IntrinsicAnything~\cite{chen2024intrinsicanything} & 13.79 & 0.67 & 0.36 & - & - & - & - & - & - & - & -\\
    IntrinsicDiff~\cite{kocsis2024intrinsic} & 17.42 & 0.80 & 0.22 & 16.46 & 0.50 & 0.29 & 13.33 & 0.57 & 0.34 & - & - \\
    RGBX~\cite{zeng2024rgb} & 18.84 & 0.81 & 0.18 & 11.92 & 0.30 & 0.41 & 11.67 & 0.51 & 0.43 & 18.57 & -
    \\ \midrule
    Ours & \textbf{21.95} & \textbf{0.87} & \textbf{0.12} & \textbf{17.64} & \textbf{0.63} & \textbf{0.22} & \textbf{16.72} &\textbf{ 0.71} & \textbf{0.24} & \textbf{12.23} & \textbf{0.08}
    \\ \bottomrule[1pt]
    \end{tabular}}
    \caption{\textbf{Quantitative comparison of inverse rendering on the synthetic InteriorVerse dataset~\cite{zhu2022learning}}. Note, ``-'' means that the prediction of a certain intrinsic property is not allowed by a method.}
    \vspace{-2mm}
    \label{table:synthetic}
\end{table*}
In contrast to most previous methods that typically focus on the estimation of single or very limited number of intrinsic properties, we aim to achieve holistic inverse rendering that jointly estimates five intrinsic properties. To this end, we use the text prompt of each intrinsic property as a trigger to adapt a single pre-trained diffusion model for prediction of multiple intrinsic properties, without modifying its architecture. Specifically, we formulate inverse rendering as a deterministic one-to-many distribution mapping problem, starting from the same latent code $Z$ and diverging along different intrinsic trajectories towards various latent intrinsics $Z_{i}$. Hence, we use the text prompt as a simple trigger~\cite{zeng2024rgb} to control the direction of the intrinsic trajectories. The training objective is defined as: 
\begin{equation}
\mathcal{L}_{flow} = \mathbb{E}_{t,Z,Z_i} \sum_{i=1}^{N}\left\| \mu_{\theta}(Z_t, t, p_{i}) - (Z_{i}-Z) \right\|_2^2,
\end{equation}
where $N$ is the number of intrinsic types, $p_{i}\in$ $\{$``albedo'', ``metallic'', ``roughness'', ``normal'', ``depth''$\}$ refers to the text trigger representing the target intrinsic property, while $Z_{i}$ is the corresponding latent intrinsic.

\vspace{0.5em}

Note, unlike previous methods that mostly use UNet for diffusion denoising, we experimentally found that Diffusion Transformer (DiT)~\cite{peebles2023scalable} is a more effective flow estimator, as its utilization of global information is crucial to the success of inverse rendering. Hence, a pre-trained DiT from Stable Diffusion V3~\cite{esser2024scaling} is fine-tuned to serve as our flow estimator $\mu_{\theta}$. Instead of fine-tuning the entire pre-trained model or adding extra ControlNet branches~\cite{zhang2023adding}, we retain the original architecture of the pre-trained DiT and apply LoRA~\cite{hu2021lora} to fine-tune a very small number of parameters.

\subsection{Generative Renderer}
Although each intrinsic property is supervised by the corresponding ground truth in the training dataset, there is actually no direct constraint between these properties and the input image, neglecting their potential relationship. To address this issue, we develop a generative renderer and accordingly design a reconstruction loss to ensure that we are able to reconstruct the input image from the predicted intrinsic properties.  Specifically, we use ControlNet~\cite{zhang2023adding} to incorporate the five intrinsic properties as the control signal to a pre-trained model from Stable Diffusion~\cite{rombach2022high} and fine-tune it as the generative renderer $\mathcal{R}$ to render the input image based on the given ground-truth intrinsic properties:
\begin{equation} 
    x_t = \sqrt{\bar{\alpha}_t} Z + \sqrt{1 - \bar{\alpha}_t} \epsilon, t \in \{0, 1, \dots, T\},
\end{equation}
\begin{equation}
\mathcal{L} = \mathbb{E}_{t,Z,\epsilon} \left\| \mathcal{R}_{\theta}(x_t, t, (A,M,R,N,D))) - \epsilon \right\|_2^2, 
\end{equation}
where $Z$ is the latent code of the input image and the condition signal is the combination of the five intrinsic properties including albedo ($A$), metallic ($M$), roughness ($R$), normal ($N$) and depth ($D$). As the scene lighting is not provided as a conditioning input, we instead
model the rendering process as property-conditioned noise-to-image mapping with stochasticity (\ie, Gaussian noise) representing the unknown lighting conditions. Specifically, the pre-trained generative renderer $\mathcal{R}$ is employed to formulate the following reconstruction loss:
\begin{equation}
\mathcal{L}_{rec} = \mathbb{E}_{t,Z,\epsilon} \left\| \mathcal{R}(\sqrt{\bar{\alpha}_t} Z+\sqrt{1 - \bar{\alpha}_t} \epsilon_t, t, \mu_\theta(I))) - \epsilon \right\|_2^2, 
\end{equation}
where $\mu_\theta(I)=(\tilde{A},\tilde{M},\tilde{R},\tilde{N},\tilde{D})$ refers to one-step estimation of all five intrinsic properties while $\epsilon_t$ is a sampled noise denoting the unknown lighting. Inspired by DreamFusion~\cite{poole2022dreamfusion}, $\mathcal{L}_{rec}$ is implemented as the Score Distillation Sampling (SDS) loss to compute the gradient for $\mu_\theta$ without backpropagating through the generative renderer $\mathcal{R}$. Specifically, we compute the gradient of \( \mathcal{L} = \mu_{\theta}(I) \cdot \text{stop\_gradient}\left[ \mathcal{L}_{rec}\right] \) with respect to the parameters $\theta$. Then, the parameters can be updated using an optimizer. 

\vspace{0.5em}
\noindent \textbf{Discussion.} Note, to remedy the lack of lighting in the reconstruction loss, we sample various different lighting in the training iterations, which is functionally equivalent to integrating all the possible lighting. Therefore, by minimizing the reconstruction loss, our model will be trained towards the direction of removing the effect of lighting and finally converges on lighting-independent intrinsic properties. Please see the supplementary material for more details about the generative renderer.

\subsection{Inference}
As shown in Figure~\ref{figure_framework}, during inference, the VAE encoder $\mathcal{E}$ encodes the input image $I$ into the latent code $Z$. Then, the directions of intrinsic trajectory are determined by specific text prompt $p$. Starting from the initial point $Z_{0}=Z$ with the prompt $p_{i}$, we can gradually transition towards the target latent intrinsic by:
\begin{equation}
Z_{t + \frac{1}{K}} = Z_t + \frac{1}{K} \mu_{\theta}(Z_t, t, p_{i}),\label{eq13}
\end{equation}
where $K$ denotes the total number of sampling steps, which is set as 10 in all our experiments. $t \in \{0, \ldots, K-1\}/K$ is the  normalized timestep. After obtaining the target latent intrinsic, we utilize the VAE decoder $\mathcal{D}$ to recover the corresponding intrinsic property.

\begin{figure*}[t]
\centering
\includegraphics[width=\textwidth]{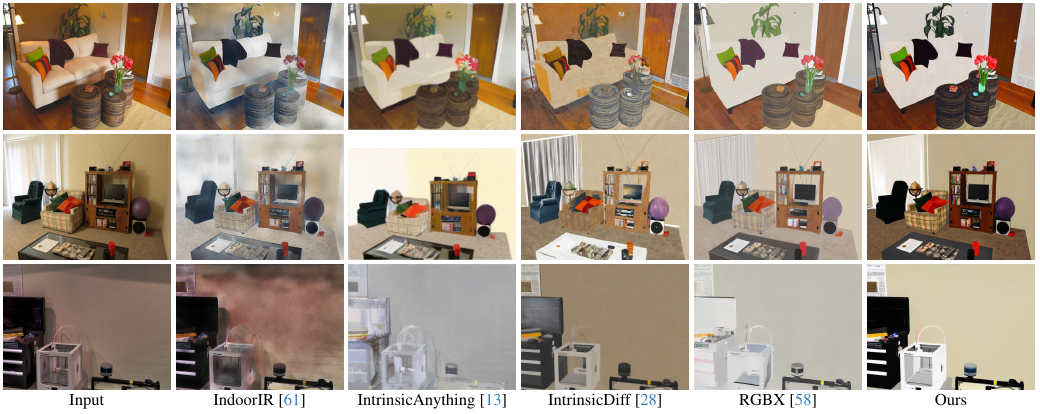} \\
\caption{\textbf{Qualitative comparison of albedo estimation on real-world images.}}
\label{figure_real_albedo}
\end{figure*}

\begin{figure*}[t]
\centering
\includegraphics[width=\textwidth]{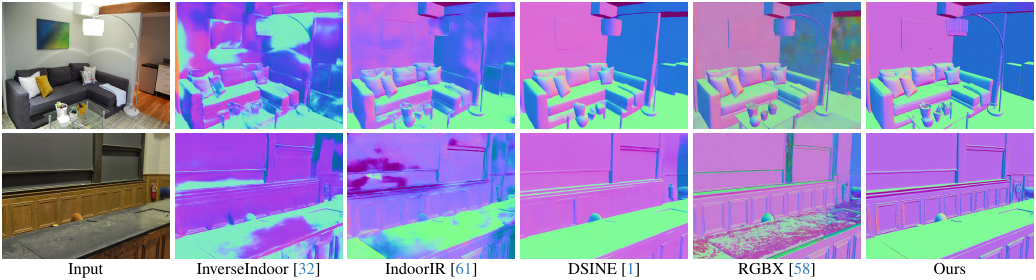} \\
\caption{\textbf{Qualitative comparison of normal estimation on real-world images}.}
\label{figure_real_wolrd_normal}
\end{figure*}

\begin{figure*}[t]
\centering
\includegraphics[width=\textwidth]{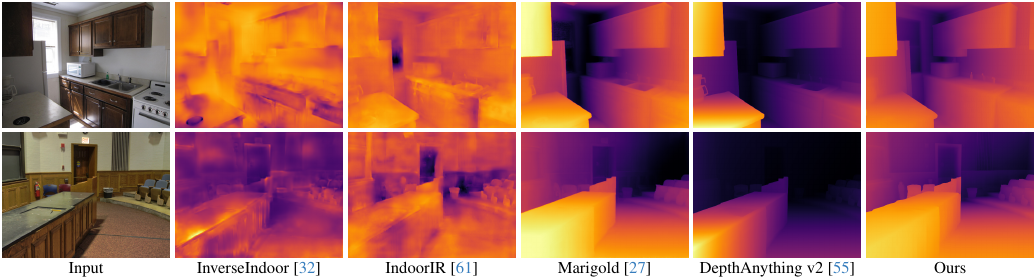} \\
\caption{\textbf{Qualitative comparison of depth estimation on real-world images}.}
\label{figure_real_wolrd_depth}
\end{figure*}

\subsection{Implementation details}
We follow previous methods~\cite{zhu2022learning,kocsis2024intrinsic} to train our model on the synthetic InteriorVerse dataset~\cite{zhu2022learning} for 100 epochs. The dataset contains ground truth for albedo, metallic, roughness, normal, and depth, having a total of 45,073 samples for training and 2,672 samples for testing. The encoder $\mathcal{E}$ and decoder $\mathcal{D}$ are from the pre-trained variational autoencoder (VAE) in Stable Diffusion V3~\cite{esser2024scaling}. During training, we apply LoRA to fine-tune the attention layers in the pre-trained diffusion transformer using the AdmaW optimizer~\cite{loshchilov2017decoupled} with a learning rate of 1e-4 and a weight decay rate of 1e-2. The rank of LoRA layer is set as 64 and the weights are initialized using Gaussian distribution. Furthermore, to adapt to pre-trained VAE, we transform the input image and intrinsic properties to the range of $[-1,1]$. For single-channel intrinsic properties, \eg, metallic, roughness, and depth, we replicate them across three channels during training to simulate the RGB image and average the output across channels during inference to recover them.

\section{Experiments}

\subsection{Evaluation Datasets and Metrics}

\qing{ We evaluate our method on several benchmark indoor datasets including the synthetic InteriorVerse dataset~\cite{zhu2022learning} and several benchmark real-world indoor datasets, \ie, IIW~\cite{bell2014intrinsic}, DIODE~\cite{DIODE}, and NYUv2~\cite{silberman2012indoor}. For quantitative evaluation of albedo, metallic, and roughness on ground-truth available synthetic datasets, we employ the commonly-used PSNR, SSIM, and LPIPS to evaluate the prediction accuracy. We follow DSINE~\cite{bae2024rethinking} to assess normal estimation by calculating the mean angular error (AE) and use the Absolute Mean Relative Error (AMRE) introduced in \cite{ke2024repurposing} for monocular depth evaluation. Note, for the IIW dataset with relative human albedo annotations rather than albedo itself, we follow \cite{bell2014intrinsic} to use Weighted Human Disagreement Rate (WHDR) to evaluate the performance of albedo prediction.
}

\begin{table}[tb]
    \centering 
    \resizebox{0.99\linewidth}{!}{
    \begin{tabular}{lccccc}
    \toprule[1pt]
    \multirow{2}{*}{Method}  & \multicolumn{1}{c}{IIW} & \multicolumn{2}{c}{NYUv2} & \multicolumn{2}{c}{DIODE} \\
    &\multicolumn{1}{c}{WHDR$\downarrow$} &\multicolumn{1}{c}{AE$\downarrow$} &\multicolumn{1}{c}{AMRE$\downarrow$}  &\multicolumn{1}{c}{AE$\downarrow$} & \multicolumn{1}{c}{AMRE$\downarrow$}  
    \\ \midrule
    InverseIndoor~\cite{li2020inverse} & 21.99 & 38.51 & 0.28  & 44.70 & 0.20  \\
    IndoorIR~\cite{zhu2022learning} & 22.90 & 34.38 & 0.23  & 36.03 & 0.19  \\
    IntrinsicAnything~\cite{chen2024intrinsicanything} & 27.75 & - & - & - & - \\
    IntrinsicDiff~\cite{kocsis2024intrinsic} & 22.02 & - & - & - & - \\
    RGBX~\cite{zeng2024rgb} & 21.80 & 35.24 & -  & 42.41 & -  
    \\ \midrule
		Ours & \textbf{20.90} & \textbf{28.40} & \textbf{0.07}  & \textbf{28.29 } & \textbf{0.11} 
    \\ \bottomrule[1pt]
    \end{tabular}}
    \caption{\textbf{Quantitative comparison of albedo and geometry estimation on real-world datasets.} For fair comparison, all the methods here are not fine-tuned on these datasets.}
    \vspace{-2mm}
    \label{table:real_com}
\end{table}
\begin{figure*}[t]
\centering
\includegraphics[width=\textwidth]{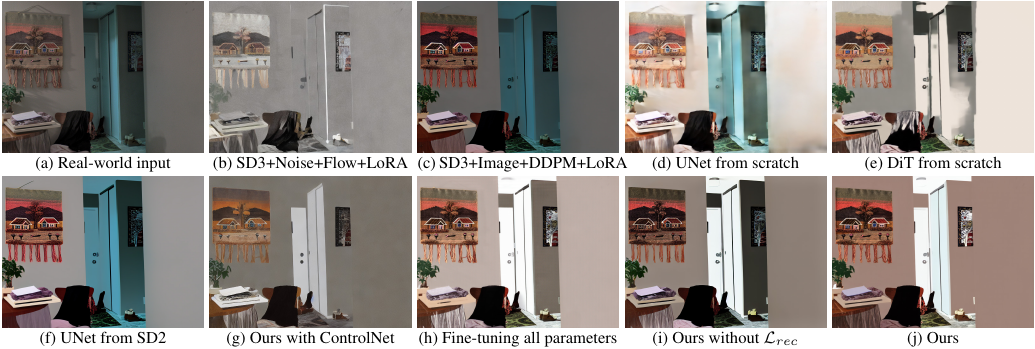} \\
\caption{\textbf{Qualitative ablation study on real-world albedo estimation.} }
\label{figure_ablation}
\end{figure*}

\subsection{Comparison with State-of-the-art Methods}
\qing{\noindent \textbf{Baselines.} We compare our method with the following state-of-the-art methods: (i) two CNN-based inverse rendering methods, InverseIndoor \cite{li2020inverse} and IndoorIR \cite{zhu2022learning}, and (ii) three latest diffusion-based methods, IntrinsicDiff \cite{kocsis2024intrinsic}, IntrinsicAnything \cite{chen2024intrinsicanything} and RGBX~\cite{zeng2024rgb}. Note, we do not compare with other existing single-image inverse rendering methods \cite{luo2024intrinsicdiffusion,li2022phyir,sengupta2019neural,wang2021learning,zhu2022irisformer}, as their codes and trained models are not publicly available. 
For fair comparisons, we produce results of the compared methods by fine-tuning their models on the InteriorVerse dataset~\cite{zhu2022learning}. }

\vspace{0.5em}
\qing{\noindent \textbf{Evaluation on synthetic data.} Table~\ref{table:synthetic} reports the quantitative comparison results of holistic inverse rendering on the synthetic InteriorVerse dataset. As shown, our method outperforms the compared methods on all five intrinsic properties. Figure~\ref{fig:syn_comp} further presents visual comparison results, where we can see that our method robustly produces high-quality predictions closer to the ground truths while avoiding the common blurring or ghosting artifacts, demonstrating the effectiveness and robustness of our method.
}

\vspace{0.5em}
\qing{\noindent \textbf{Evaluation on real-world data.} In addition to synthetic data, we also evaluate our method on real-world data. Specifically, we evaluate albedo estimation on the IIW dataset and conduct normal and depth comparisons on the NYUv2 and DIODE datasets. As shown in Table~\ref{table:real_com}, our method exhibits a clear advantage over the compared methods on these real-world datasets. Figure~\ref{figure_real_albedo} presents visual comparisons on albedo estimation, where we can see that our method produces albedo predictions that are more faithful to the input images. Visual comparisons on normal and depth estimation are given in Figures~\ref{figure_real_wolrd_normal} and \ref{figure_real_wolrd_depth}, respectively. As can be seen, despite that our method is fine-tuned on the synthetic InteriorVerse dataset with only 45K training images, it can produce very competitive normal and depth prediction results, which are comparable to current leading methods (basically trained on datasets significantly larger than ours) specifically designed for normal \cite{bae2024rethinking} and depth \cite{depth_anything_v2,ke2024repurposing} estimation. Please see the supplementary material for more results as well as comparisons on roughness and metallic estimation.
}

\begin{table}
	\centering
	\vspace{-2mm}
    \resizebox{\linewidth}{!}{
	\begin{tabular}{lcccc}
		\toprule[1pt]
		Method & \multicolumn{1}{l}{Params} & \multicolumn{1}{l}{PSNR$\uparrow$} & \multicolumn{1}{l}{SSIM$\uparrow$} & \multicolumn{1}{l}{LPIPS$\downarrow$} 
    \\ \midrule
        SD3+Noise+Flow+LoRA & 19M & 18.25 & 0.73 & 0.22 \\
        SD3+Image+DDPM+LoRA & 19M & 17.62 & 0.81 & 0.18 \\
    \midrule
        UNet from scratch & 100M & 17.99 & 0.78 & 0.25 \\
        DiT from scratch & 300M & 18.80 & 0.81 & 0.17 \\ 
        UNet from SD2 & 13M & 19.51 & 0.83 & 0.15 \\
    \midrule
        Ours with ControlNet & 1000M & 17.13 & 0.68 & 0.27 \\
        Ours fine-tuning all parameters& 2047M & 21.30 & 0.85 & 0.14 \\
        Ours w/o reconstruction loss $\mathcal{L}_{rec}$ & 19M & 21.03 & 0.82 & 0.14 \\ \midrule
        Ours (SD3+Image+Flow+LoRA) & 19M &\textbf{21.95} & \textbf{0.87} & \textbf{0.12}
    \\ \bottomrule[1pt]
	\end{tabular}}
        \caption{\textbf{Quantitative ablation study on albedo estimation on the InteriorVerse dataset~\cite{zhu2022learning}.} }
	\vspace{-2mm}
	\label{table:ablation}
\end{table}
\begin{figure}[t]
\centering
\includegraphics[width=\linewidth]{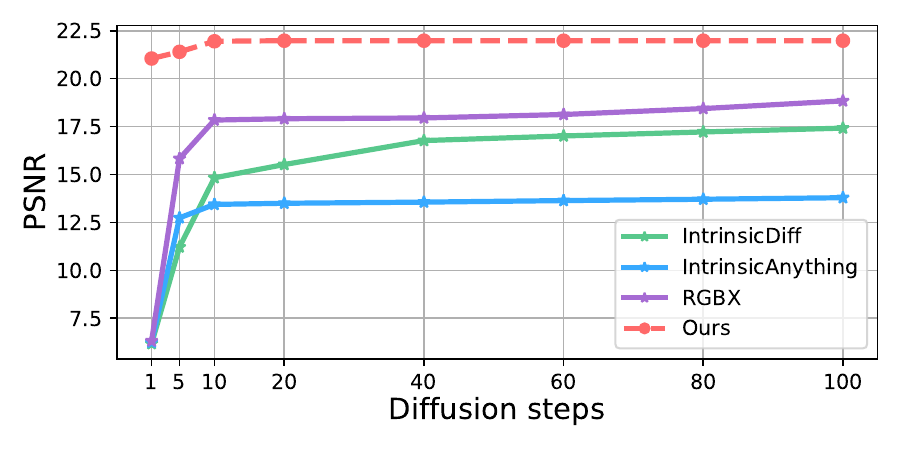} \\
\vspace{-4mm}
\caption{\textbf{Quantitative comparison of albedo estimation using varying diffusion steps on the InteriorVerse dataset.} In comparison to recent diffusion-based inverse rendering methods, \eg, IntrinsicDiff~\cite{kocsis2024intrinsic}, IntrinsicAnything~\cite{chen2024intrinsicanything}, and RGBX~\cite{zeng2024rgb}, our method achieves better albedo estimation in fewer diffusion steps.}
\label{figure_steps}
\end{figure}
\begin{figure}[t]
\centering
\includegraphics[width=\linewidth]{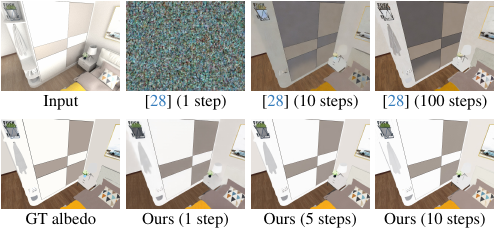} \\
\caption{\textbf{Qualitative comparison of albedo estimation using different number of diffusion steps.} }
\label{figure_timestep}
\end{figure}

\subsection{Ablation Study}
\noindent \textbf{Effect of deterministic image-to-intrinsic mapping.} To examine the effectiveness of our image-to-intrinsic mapping design, we compare it with several alternative designs: (i) \textbf{Noise+Flow+LoRA}: image-conditioned noise-to-intrinsic mapping learned by flow matching while fine-tuning with LoRA. (ii) \textbf{Image+DDPM+LoRA}: image-to-intrinsic mapping using v-prediction of DDPM while fine-tuning with LoRA. As shown in Figure~\ref{figure_ablation}(b)-(c) and Table~\ref{table:ablation} (see 2nd and 3rd rows), our approach achieves better results than the alternatives, validating the superiority of our deterministic image-to-intrinsic mapping design.

\vspace{0.5em}
\noindent \textbf{Effect of diffusion prior.} To evaluate the effectiveness of the pre-trained diffusion prior, we compare our method with two alternative models trained from scratch with UNet and diffusion transformer (DiT), respectively. As shown in Figure \ref{figure_ablation}(d)-(e) and Table~\ref{table:ablation} (see 4th and 5th rows), despite having more trainable parameters, the two models trained from scratch fail to produce satisfactory results, verifying the effectiveness of the diffusion prior.

\vspace{0.5em}
\noindent \textbf{Effect of Diffusion Transformer (DiT).} Figure \ref{figure_ablation}(f) and the 6th row in Table~\ref{table:ablation} provide a comparison between UNet from Stable Diffusion v2 (SD2) and our utilized DiT from Stable Diffusion v3 (SD3). As shown, replacing DiT with UNet leads to an obvious performance drop.

\vspace{0.5em}
\noindent \textbf{Effect of different fine-tuning strategies.} Figure \ref{figure_ablation}(g)-(h) and Table~\ref{table:ablation} (see 7th and 8th rows) provide comparisons between fine-tuning with ControlNet, fine-tuning all parameters, and fine-tuning with LoRA (ours). As shown, our fine-tuning with LoRA strategy achieves the best performance while corresponding to the fewest trainable parameters.

\vspace{0.5em}
\noindent \textbf{Effect of the reconstruction loss $\mathcal{L}_{rec}.$} As shown in the 9th row in Table~\ref{table:ablation}, the reconstruction loss $\mathcal{L}_{rec}$ is beneficial to obtain better results. Moreover, Figure~\ref{figure_ablation}(i) shows that omitting the loss would lead to an albedo prediction with clear shadow residuals on the wall.

\begin{figure}[t]
\centering
\includegraphics[width=\linewidth]{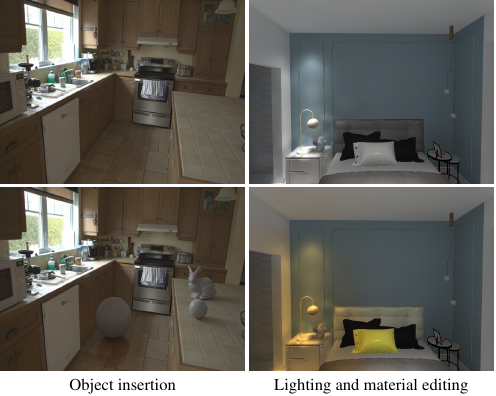} \\
\caption{\textbf{Applications. } Our method allows to produce photorealistic object insertion, as well as lighting and material editing. }
\label{figure_object_insert}
\end{figure}

\vspace{0.5em}
\qing{\noindent \textbf{Effect of different number of diffusion steps.} We evaluate the performance of our method with different numbers of diffusion steps in Figures \ref{figure_steps} and \ref{figure_timestep}. In contrast to the latest diffusion-based methods \cite{kocsis2024intrinsic,chen2024intrinsicanything,zeng2024rgb}, our approach achieves superior results using significantly fewer steps.}

\subsection{Applications}
Thanks to the high-quality material and geometry properties recovered from a single image, our method enables various applications. As shown in Figure \ref{figure_object_insert}, our method can produce visually pleasing results for object insertion, lighting, and material editing. Note, although our method itself does not allow for lighting estimation because our training dataset does not provide ground-truth for lighting, we can estimate it via image reconstruction optimization. Specifically, we use pre-integrated environment lighting parameterized by Spherical Gaussians (SG) for global illumination, along with 48 SG emission profiles to represent point lights~\cite{kocsis2024intrinsic}. The SG parameters are then determined by optimizing a reconstruction loss between the re-rendered output and the input. With the estimated lighting, we can achieve controllable lighting editing and 3D object insertion. 

\section{Conclusion}
We have presented DNF-Intrinsic, a deterministic noise-free diffusion-based approach that allows for robust yet efficient single-image inverse rendering of indoor scenes. Unlike previous methods that learn image-conditioned noise-to-intrinsic mapping, we propose to take the source image rather than Gaussian noise as input to learn deterministic image-to-intrinsic mapping via flow matching. By this means, our method achieves more robust intrinsic prediction performance by taking full advantage of the visual information in the source image while avoiding the high computational cost from multiple denoising steps. A generative renderer is also designed to ensure the resulting intrinsic properties are physically meaningful. Experiments show that our method outperforms previous approaches.
 
\vspace{0.5em}
\noindent \textbf{Acknowledgement.} This work was supported by the National Natural Science Foundation of China (62471499), the Guangdong Basic and Applied Basic Research Foundation (2023A1515030002).
{
    \small
    \bibliographystyle{ieeenat_fullname}
    \bibliography{main}
}
\clearpage \appendix 
\section{Details of Generative Renderer} 
\begin{figure}[t]
\centering
\includegraphics[width=\linewidth]{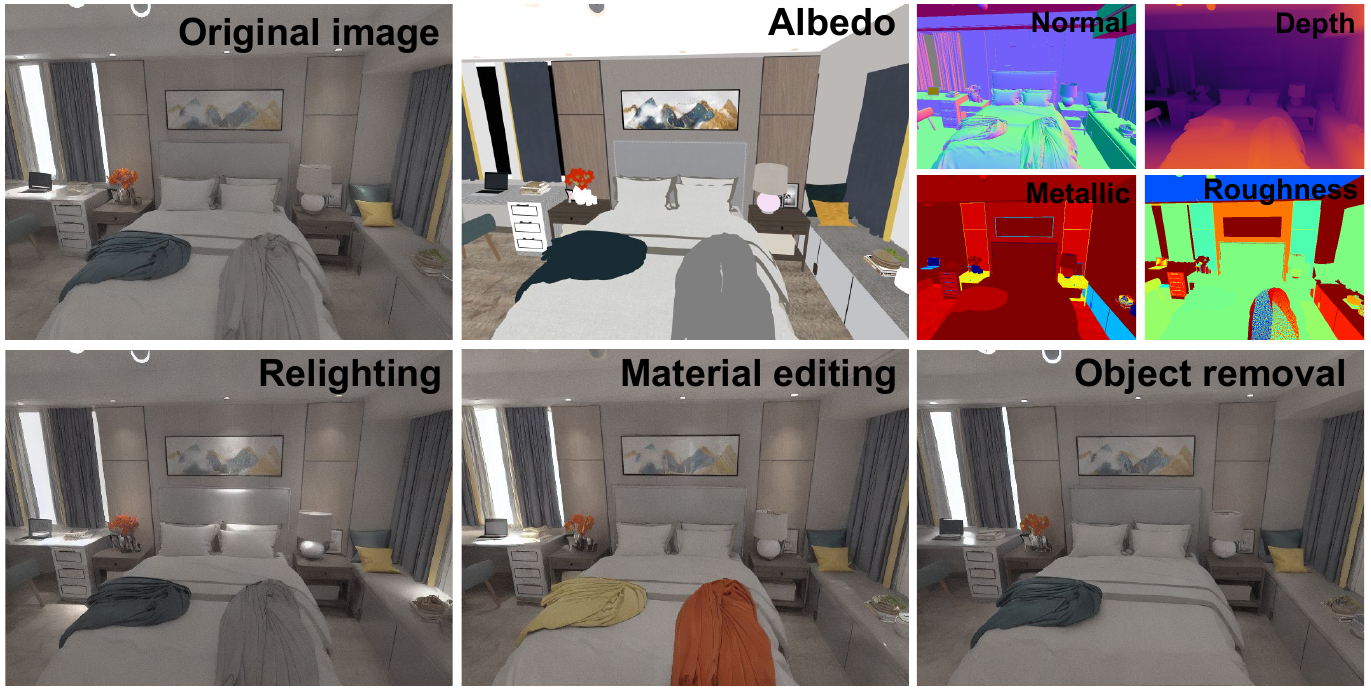} \\
\vspace{-2mm}
\caption{Examples of our generative renderer.}
\vspace{-5mm}
\label{figure_generative_renderer}
\end{figure}

Our generative renderer aims to take the scene's intrinsic properties as input and generate the input image as output. It is fine-tuned from Stable Diffusion v2, with a trainable ControlNet conditioned on 9-channel intrinsic properties (albedo, metallic, roughness, normal, and depth). The model is optimized by the AdamW optimizer with a learning rate of 1e-4 and a weight decay rate of 1e-2. Unlike traditional physics-based renderers, our generative renderer does not require environmental lighting as input and can still produce realistic rendered images. Besides, as shown in Figure~\ref{figure_generative_renderer}, our generative renderer enables various potential applications, including uncontrollable relighting, material editing, and object removal. Specifically, given the scene's intrinsic properties, our generative renderer can generate images with lighting conditions different from the original image, although this is uncontrollable since the lighting is represented by Gaussian noise. Additionally, we can manually adjust the albedo color for material editing or remove the target object from the intrinsic properties for object removal.

\section{Details of Reconstruction Loss}
Similar to Dreamfusion~\cite{poole2022dreamfusion}, we first compute the gradient of \( \mathcal{L} = \mu_{\theta}(I) \cdot \text{stop\_gradient}\left[ \mathcal{L}_{rec}\right] \) with respect to the parameters $\theta$ of the inverse rendering model, where $\mu_{\theta}(I)$ denotes the predicted intrinsic property and $I$ is the input image. Then, the parameters can be updated using an optimizer. The pseudo-code is shown in Figure~\ref{fig:score_sampling_code}.

\section{Details of Application}
In order to achieve controllable lighting editing or object insertion, we provide a solution to optimize the environmental lighting based on the high-quality intrinsic properties predicted by our method. Specifically, similar to ~\cite{kocsis2024intrinsic}, we use pre-integrated environment lighting parameterized by Spherical Gaussians (SG) for global illumination, along with 48 SG emission profiles to represent point lights. The SG parameters are then determined by optimizing a L2 loss between the re-rendered output and the input. After fitting, the parameters of the light sources,  such as color or intensity, can be adjusted independently to achieve controllable lighting editing. Meanwhile, with the environmental lighting of the scene, we can render an image with the inserted 3D object using the estimated lighting to achieve realistic object insertion.

\section{More Visual Comparison on Synthetic Data}
\begin{figure*}[t]
\centering
\includegraphics[width=\textwidth]{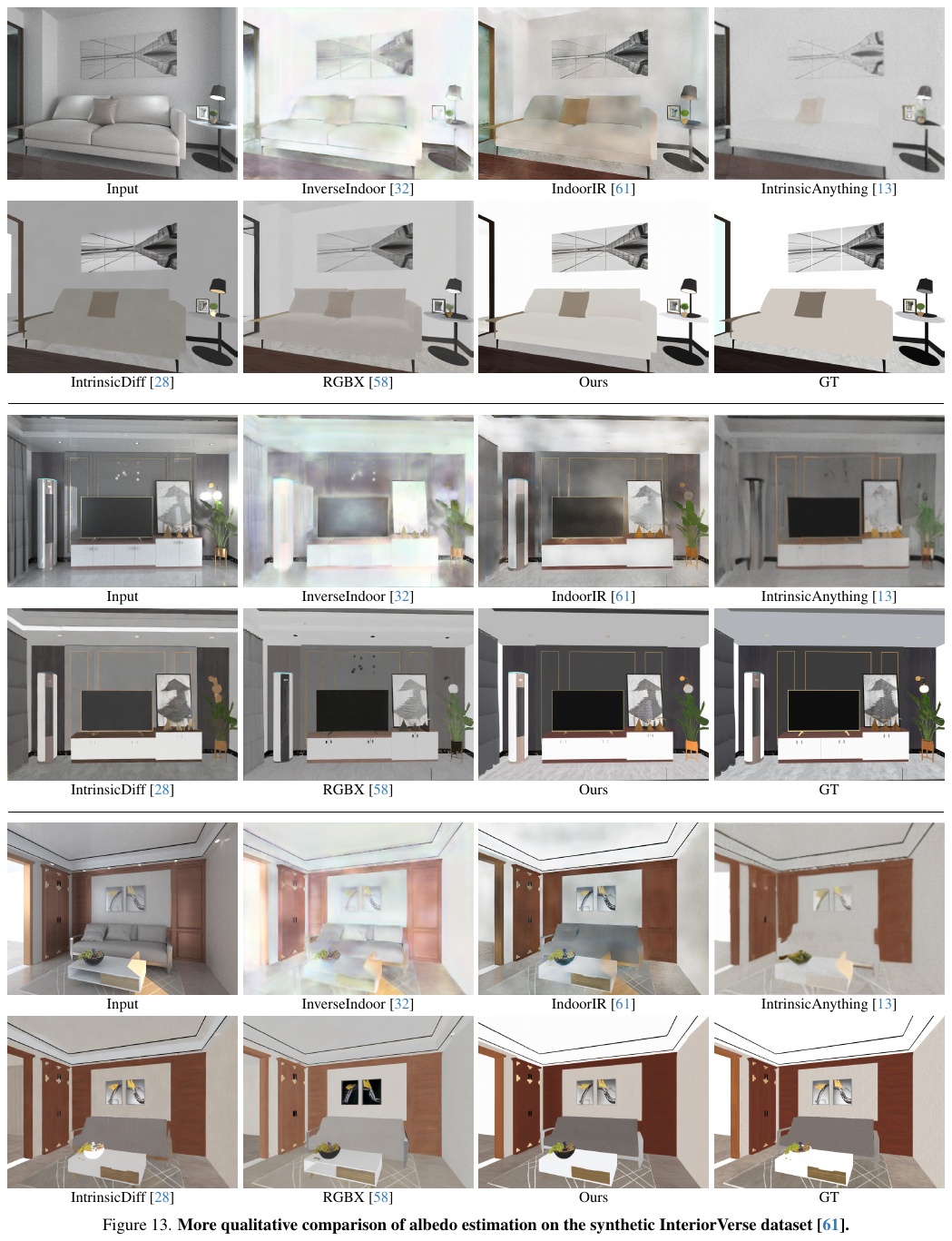} \\
\caption{\textbf{More qualitative comparison of albedo estimation on the synthetic InteriorVerse dataset~\cite{zhu2022learning}.}}
\label{figure_syn_albedo_1}
\end{figure*}

\begin{figure*}[t]
\centering
\vspace{-4mm}
\includegraphics[width=\textwidth]{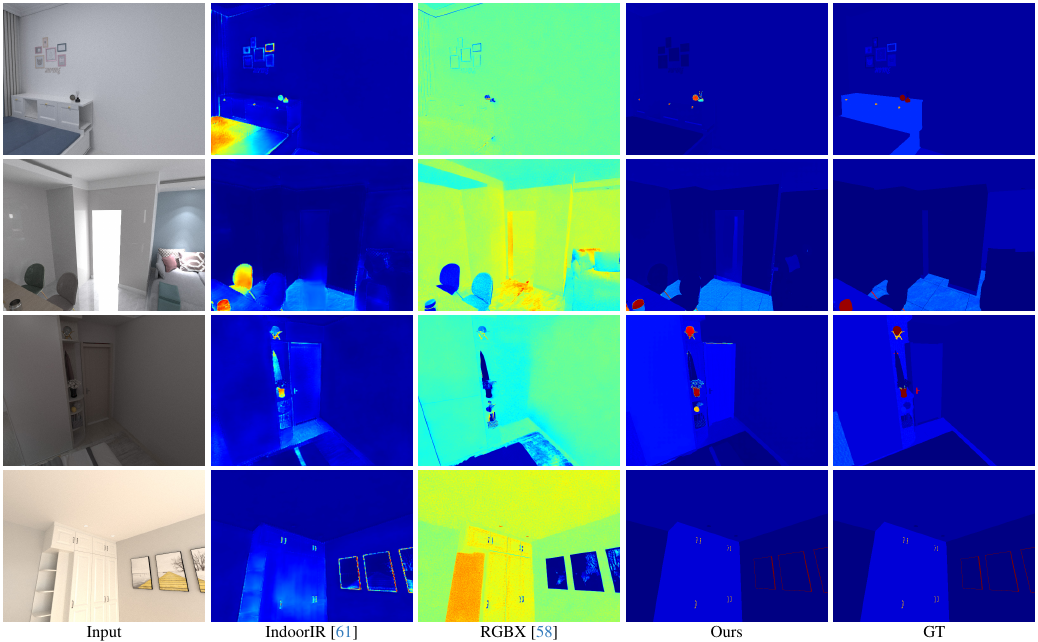} \\
\vspace{-2mm}
\caption{\textbf{More qualitative comparison of metallic estimation on the synthetic InteriorVerse dataset~\cite{zhu2022learning}.}}
\label{figure_syn_metal}
\end{figure*}

\begin{figure*}[t]
\centering
\vspace{-4mm}
\includegraphics[width=\textwidth]{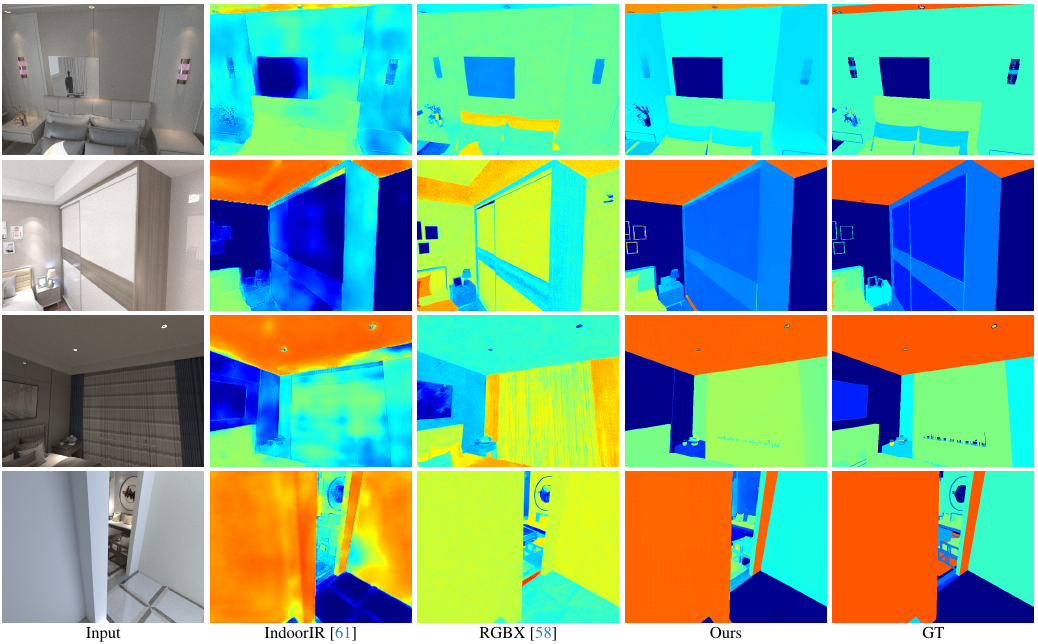} \\
\vspace{-2mm}
\caption{\textbf{More qualitative comparison of roughness estimation on the synthetic InteriorVerse dataset~\cite{zhu2022learning}.}}
\label{figure_syn_Rough}
\end{figure*}

\begin{figure*}[t]
\centering
\includegraphics[width=\textwidth]{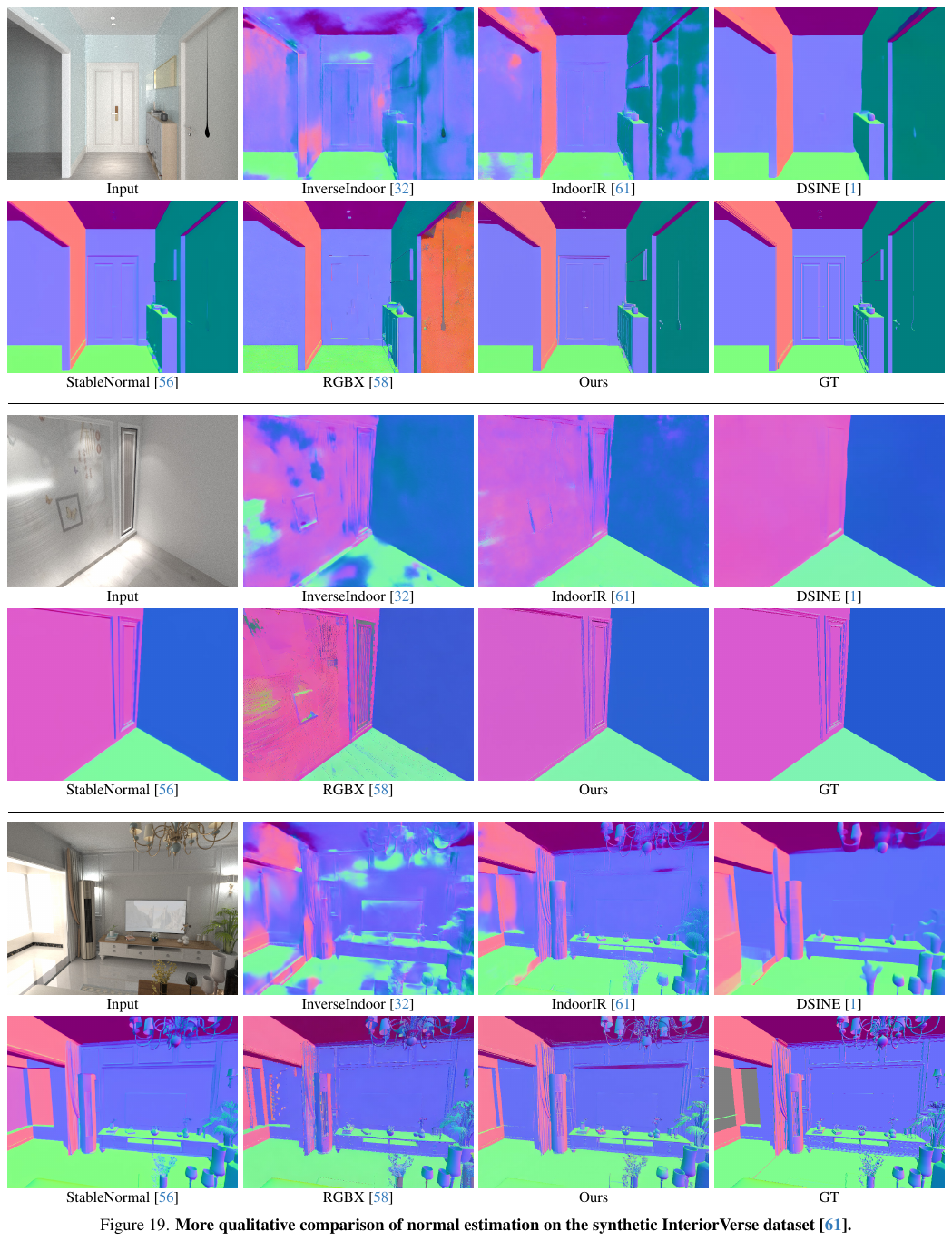} \\
\caption{\textbf{More qualitative comparison of normal estimation on the synthetic InteriorVerse dataset~\cite{zhu2022learning}.}}
\label{figure_syn_normal}
\end{figure*}

\begin{figure*}[t]
\centering
\includegraphics[width=\textwidth]{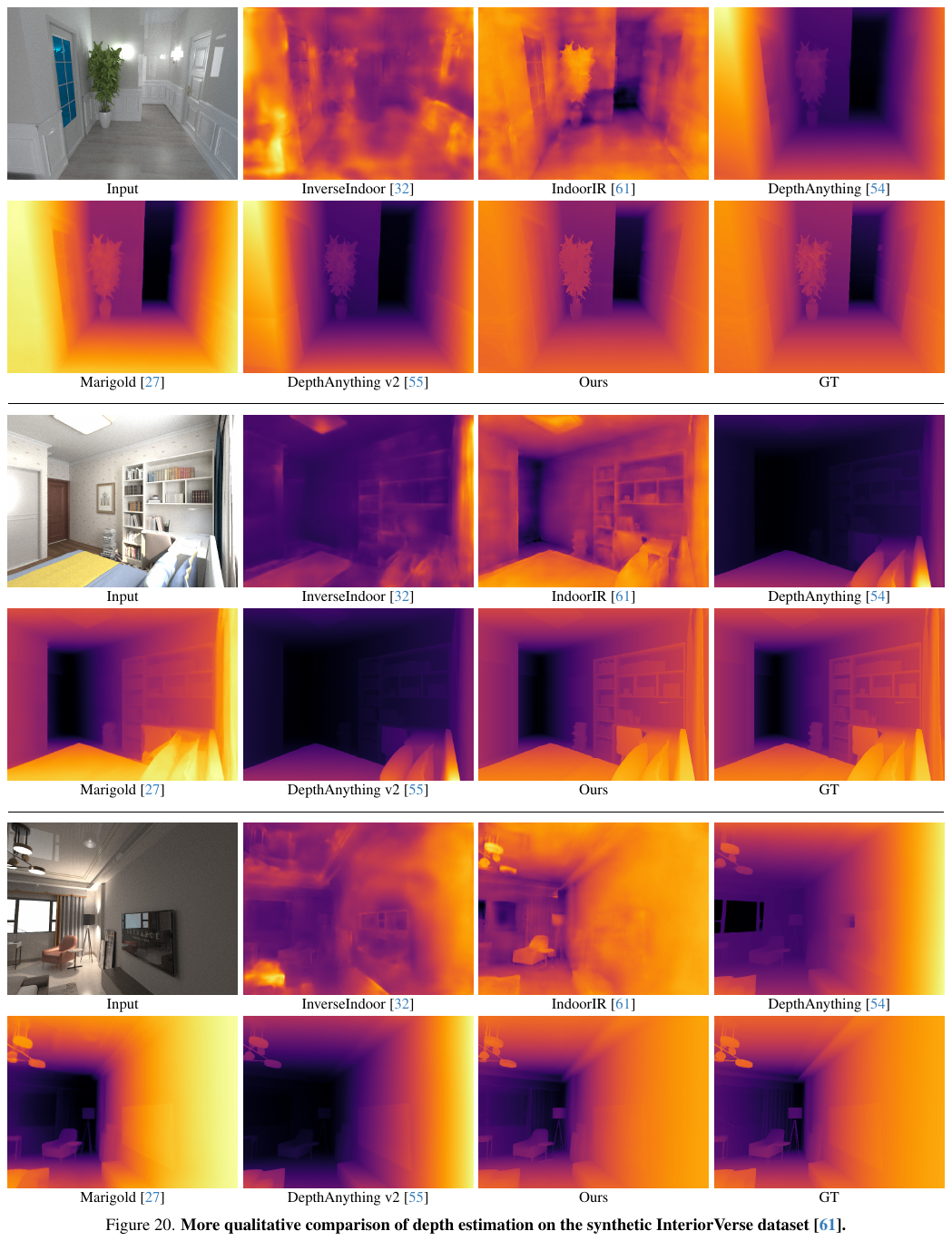} \\
\caption{\textbf{More qualitative comparison of depth estimation on the synthetic InteriorVerse dataset~\cite{zhu2022learning}.}}
\label{figure_syn_depth}
\end{figure*}

Figures~\ref{figure_syn_albedo_1}, \ref{figure_syn_metal}, \ref{figure_syn_Rough}, \ref{figure_syn_normal}, and \ref{figure_syn_depth} provide more visual comparison of inverse rendering on the synthetic InteriorVerse dataset~\cite{zhu2022learning}. As shown, our method clearly outperforms previous methods on material and geometry estimation.

\section{More Visual Comparison on Real Data}
\begin{figure*}[t]
\centering
\includegraphics[width=\textwidth]{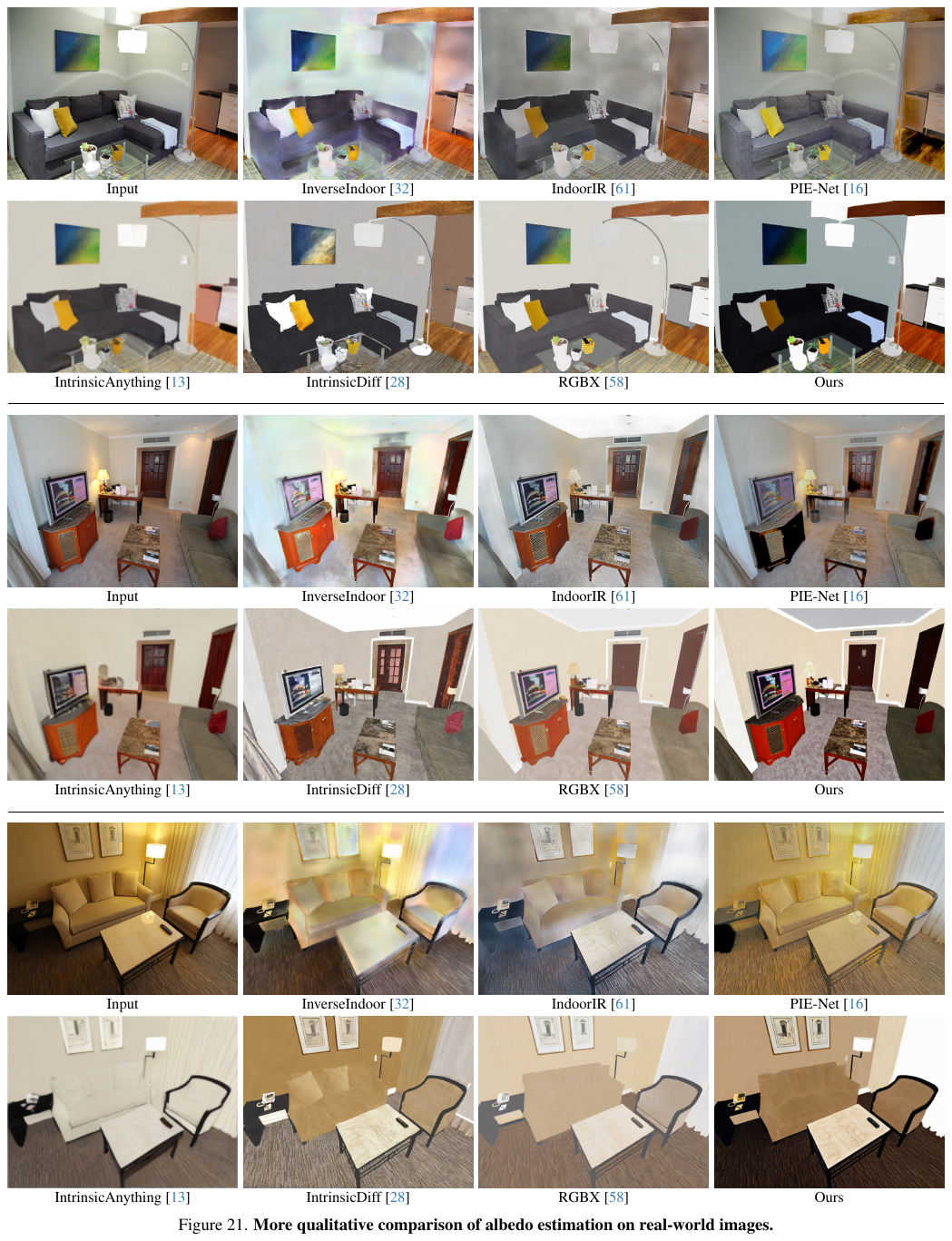} \\
\caption{\textbf{More qualitative comparison of albedo estimation on real-world images.}}
\label{figure_real_albedo_1}
\end{figure*}

\begin{figure*}[t]
\centering
\vspace{-4mm}
\includegraphics[width=\textwidth]{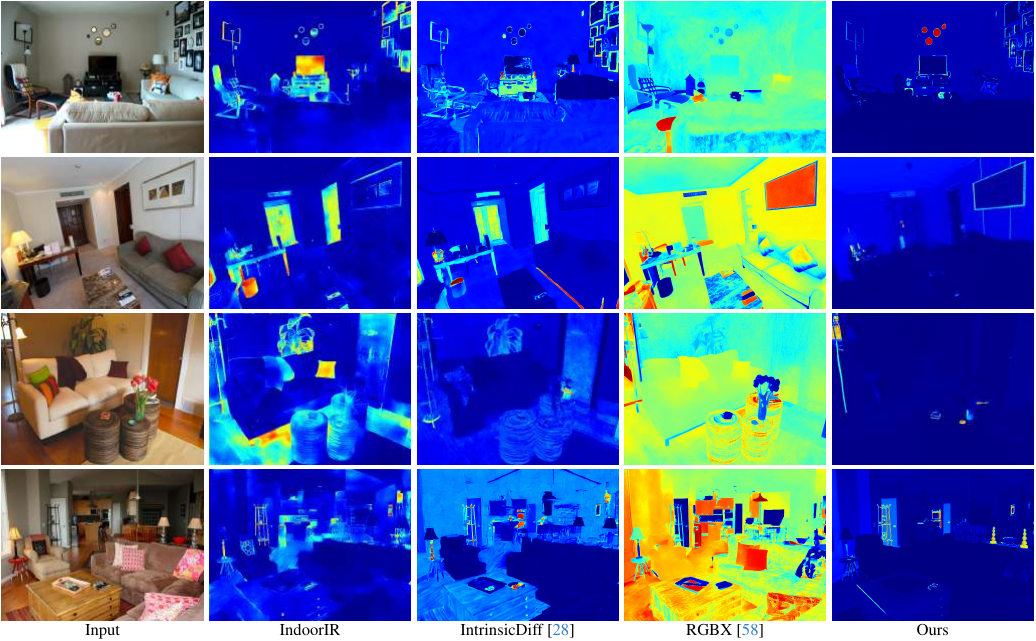} \\
\vspace{-2mm}
\caption{\textbf{More qualitative comparison of metallic estimation on real-world images.}}
\label{figure_real_metal}
\end{figure*}

\begin{figure*}[t]
\centering
\vspace{-4mm}
\includegraphics[width=\textwidth]{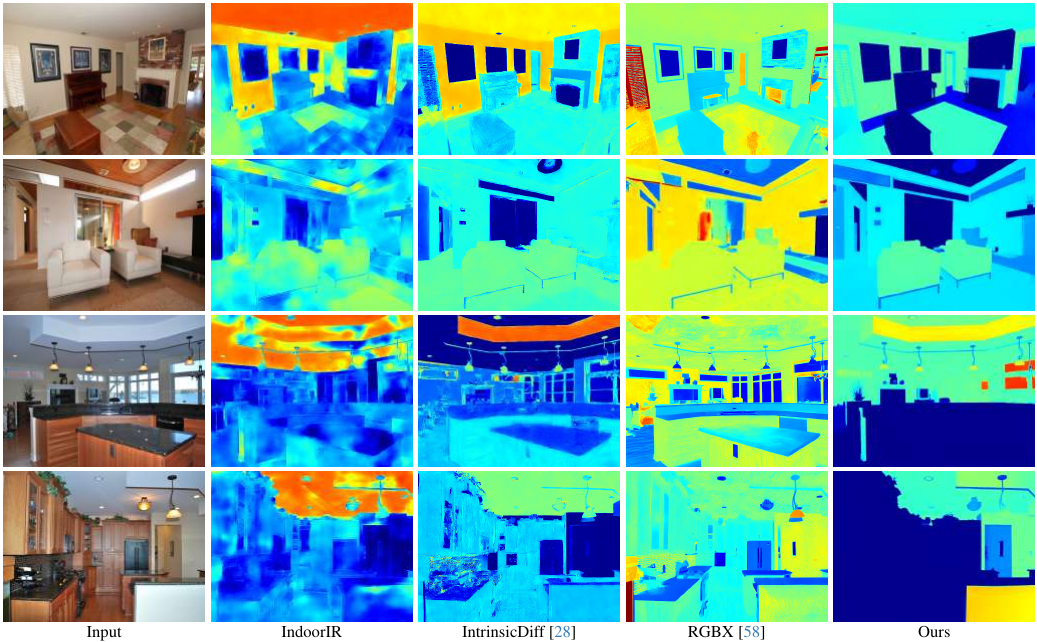} \\
\vspace{-2mm}
\caption{\textbf{More qualitative comparison of roughness estimation on real-world images.}}
\label{figure_real_rough}
\end{figure*}

\begin{figure*}[t]
\centering
\includegraphics[width=\textwidth]{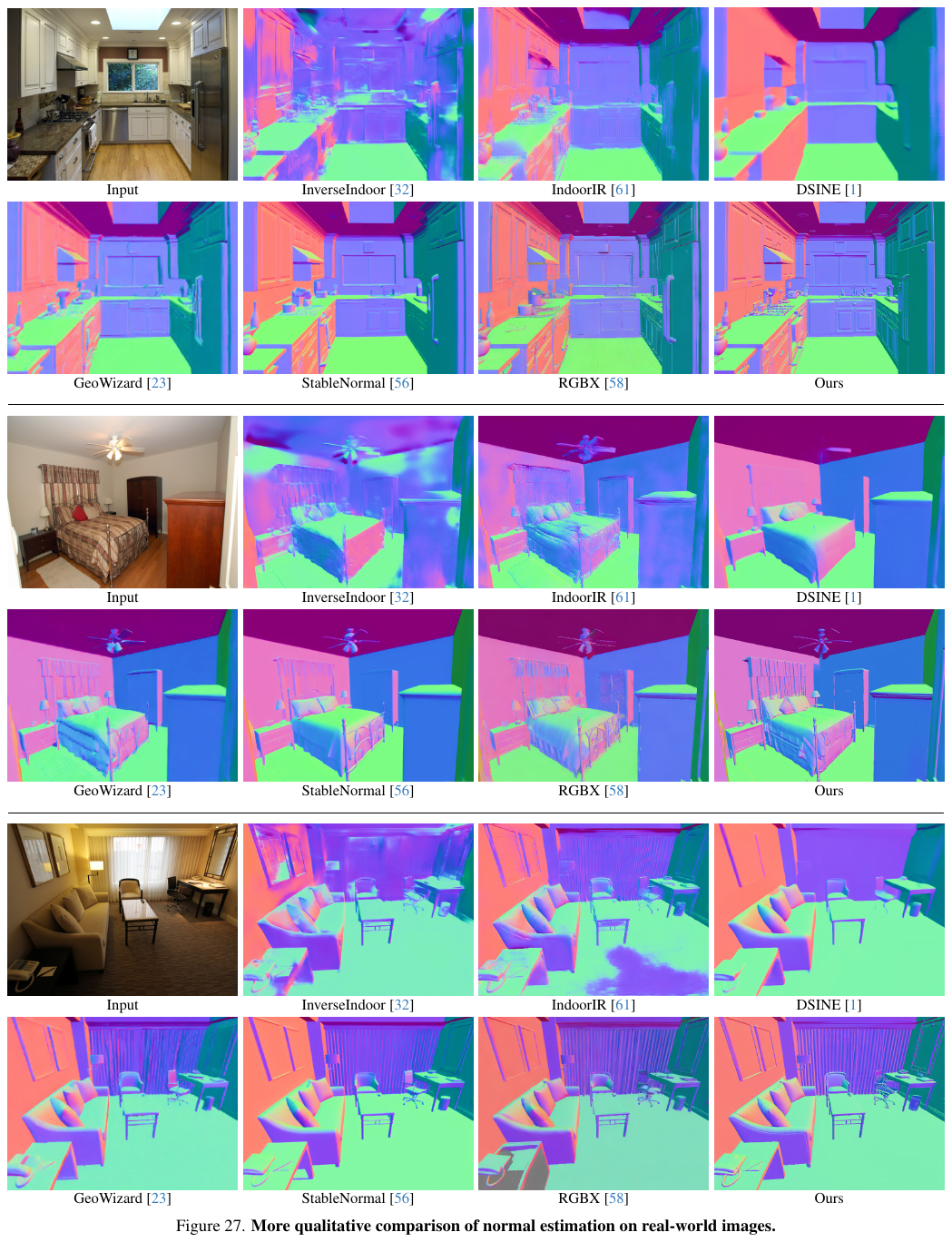} \\
\caption{\textbf{More qualitative comparison of normal estimation on real-world images.}}
\label{figure_real_normal}
\end{figure*}

\begin{figure*}[t]
\centering
\includegraphics[width=\textwidth]{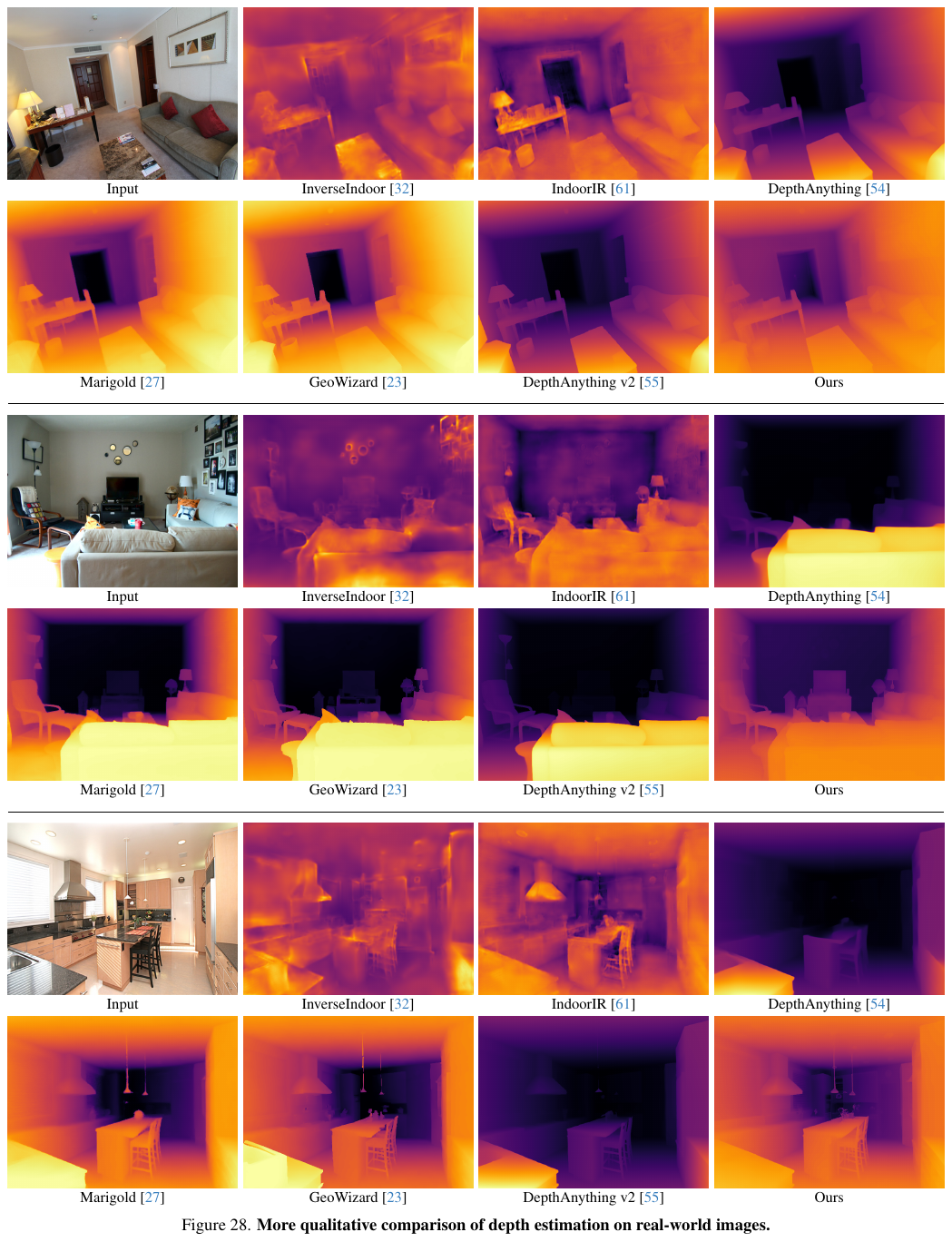} \\
\caption{\textbf{More qualitative comparison of depth estimation on real-world images.}}
\label{figure_real_depth}
\end{figure*}

Figures~\ref{figure_real_albedo_1},  \ref{figure_real_metal}, \ref{figure_real_rough}, \ref{figure_real_normal}, and \ref{figure_real_depth} provide more visual comparison on material (albedo, metallic, and roughness) and geometry estimation (depth and normal). Comparing the results, it is clear that our method outperforms current state-of-the-art inverse rendering methods, and is able to produce comparable or even better results than specialized methods for material and geometry estimation. 

\section{More Application Results}
\begin{figure*}[t]
\centering
\vspace{-4mm}
\includegraphics[width=\textwidth]{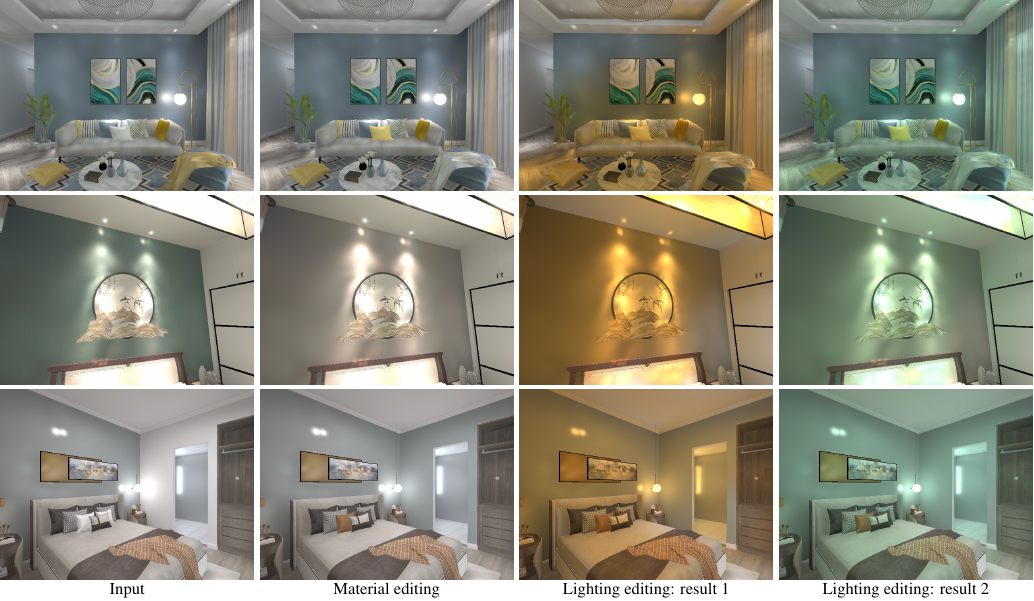} \\
\vspace{-2mm}
\caption{\textbf{More results of material and lighting editing.}}
\label{figure_lighting_editing}
\end{figure*}

We in Figure~\ref{figure_object_insert} provide more virtual object insertion results, while Figure~\ref{figure_lighting_editing} gives additional results on material and lighting editing. As shown, these application results are visually natural, manifesting the robustness of our predicted intrinsic properties. 

\begin{figure*}[t]
\begin{lstlisting}[language=Python]
params = IR_model.init() # inverse rendering model 
opt_state = optimizer.init(params)
generative_renderer = diffusion.load_ControlNet()
for nstep in iterations:
  t = random.uniform(0., 1.)
  alpha_t, sigma_t = diffusion.get_coeffs(t)
  eps = random.normal(img_shape) # sample a noise from Gaussian distribution, representing the unknown lighting
  intrinsics = IR_model(input_image)  # Get an one-step intrinsic properties observation.
  x = input_image
  z_t = alpha_t * x + sigma_t * eps  # Diffuse observation.
  epshat_t = generative_renderer.epshat(z_t, intrinsics, t)  # Score function evaluation.
  L_rec = epshat_t - eps # generative reconstuction loss
  g = grad(dot(stopgradient[L_rec], intrinsics), params)
  params, opt_state = optimizer.update(g, opt_state)  # Update params with optimizer.
return params
\end{lstlisting}
\caption{Pseudo code for the SDS-based reconstruction loss via the generative renderer that defines a differentiable mapping from parameters to intrinsic properties.The gradient \texttt{g} is computed without backpropagating through the generative renderer's U-Net. We used the \texttt{stopgradient} operator to express the loss, but the gradient of the parameter can also be easily computed as: {\texttt{g = matmul($L_{rec}$, grad(intrinsics, params))}}.}
\label{fig:score_sampling_code}
\end{figure*}


\begin{figure*}[t]
\centering
\vspace{-4mm}
\includegraphics[width=\textwidth]{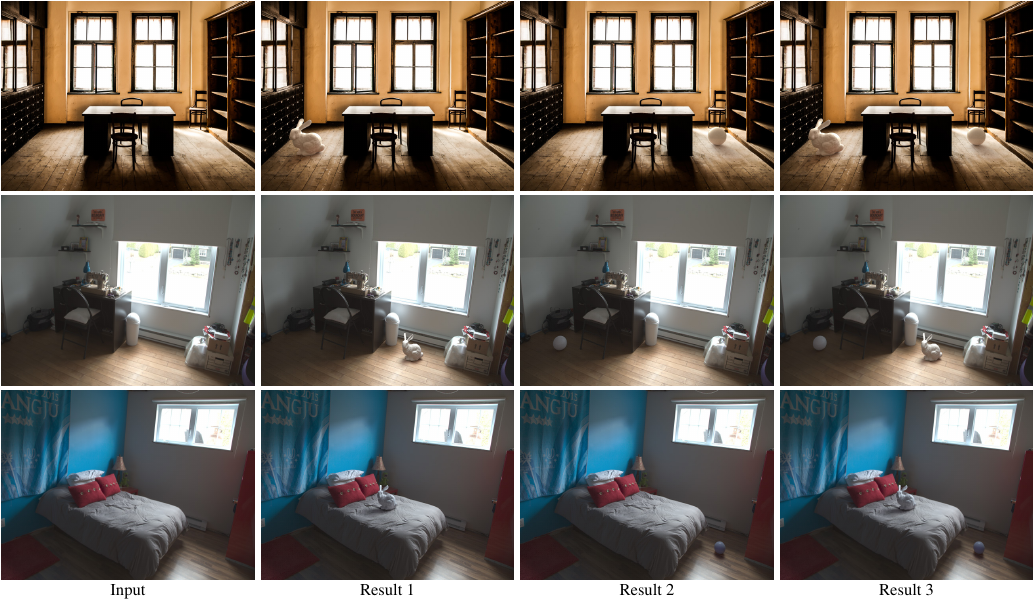} \\
\vspace{-2mm}
\caption{\textbf{More results of virtual object insertion.}}
\label{figure_object_insert}
\end{figure*}

\end{document}